# A PRELIMINARY SURVEY ON OPTIMIZED MULTIOBJECTIVE METAHEURISTIC METHODS FOR DATA CLUSTERING USING EVOLUTIONARY APPROACHES

Ramachandra Rao Kurada[1], Dr. K Karteeka Pavan[2], and Dr. AV Dattareya Rao[3]

[1]Research Scholar (Part-time), Acharya Nagarjuna University, Guntur & Assistant Professor Sr., Department of Computer Applications, Shri Vishnu Engineering College for Women, Bhimavaram
[2]Professor, Department of Information Technology, RVR & JC College of Engineering, Guntur
[3]Principal, Acharya Nagarjuna University, Guntur

## ABSTRACT

*The present survey provides the state-of-the-art of research, copiously devoted to Evolutionary Approach (EAs) for clustering exemplified with a diversity of evolutionary computations. The Survey provides a nomenclature that highlights some aspects that are very important in the context of evolutionary data clustering. The paper missions the clustering trade-offs branched out with wide-ranging Multi Objective Evolutionary Approaches (MOEAs) methods. Finally, this study addresses the potential challenges of MOEA design and data clustering, along with conclusions and recommendations for novice and researchers by positioning most promising paths of future research.*

*MOEAs have substantial success across a variety of MOP applications, from pedagogical multifunction optimization to real-world engineering design. The survey paper noticeably organizes the developments witnessed in the past three decades for EAs based metaheuristics to solve multiobjective optimization problems (MOP) and to derive significant progression in ruling high quality elucidations in a single run. Data clustering is an exigent task, whose intricacy is caused by a lack of unique and precise definition of a cluster. The discrete optimization problem uses the cluster space to derive a solution for Multiobjective data clustering. Discovery of a majority or all of the clusters (of illogical shapes) present in the data is a long-standing goal of unsupervised predictive learning problems or exploratory pattern analysis.*

## 1. INTRODUCTION

The incipient Millennium witnesses the materialization of Information Technology because the thrust behind the encroachment in Computational Intelligence (CI) [1]. The growing complexity of computer programs, availability, increased speed of computations and their ever-decreasing costs have already manifested a momentous impact on CI. Amongst the computational paradigms, Evolutionary Computation (EC) [2] is currently apperceived as a notably applicable sundry of ancient and novel computational applications in data clustering. EC comprises a set of soft-computing paradigms [3] designed to solve optimization problems. In contrast with the rigid/static models of hard computing, these nature-inspired models provide self-adaptation mechanisms, which aim at identifying and exploiting the properties of the instance of the problem being solved.

Clustering is the important step for many errands in machine learning [70]. Every algorithmic rule has its own bias attributable to the improvements of various criteria. Unsupervised machine learning is inherently an optimization task; one is trying to fit the best model to a sample of data [4]. The definition of "best" is unconditional; generalization concert with relevancy to the full universe of data points. However machine learning algorithms do not understand this a priori, and rather than rely on heuristic estimations considering the standards of their replica and restriction, such as goodness of fit with relevance to the experiment facts and figures, model parsimony, and so on [5]. Optimization [6] is that the method of getting the simplest result or profit beneath a given set of attenuating factors. Enterprise conclusions were tailored ultimately by maximizing/minimizing a goal or advantage. The dimensions and complexity of improvement issues that can be explained in a reasonable time has been advanced by the advent of up to date computing technologies.

The single-criterion optimization problem has a single optimization solution with a single objective function. In a multi-criterion optimization firm, there is more than one objective function, each of which may have an uncooperative self-optimal decision. These optimal solutions follow the name of an economist Vilfredo Pareto [7], stated in 1896 a concept known as "Pareto optimality". In a multi-criterion optimization firm, there is more than one objective function, each of which may have an uncooperative self-optimal decision. The concept is that the solution to a multi-objective optimization quandary is mundanely not a single value but instead of a set of values, withal called the "Pareto Set". A number of Pareto optimal solutions can, in principle, is captured in a EAs population, thereby sanctioning a utilizer to find multiple Pareto-optimal solutions in one simulation.

In topical times MOPs are the crucial areas in science and engineering. The complexity of MOPs becomes more and more significant in terms of size of the problem to be solved i.e. the number of objective functions and size of the search space [8]. The taxonomy of MOP is shown in Figure 1.

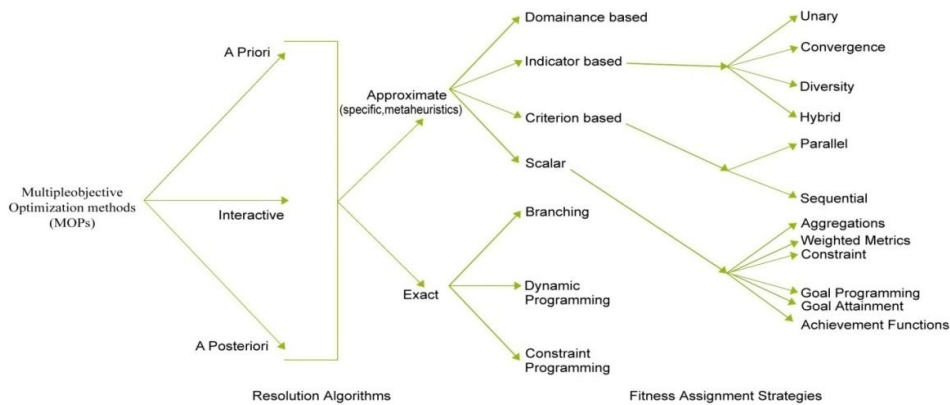

Figure 1. Taxonomy of Multiobjective Optimization Problems (MOPs)

MOPs do multi criteria decision making by three approaches Apriori, A posteriori and Interactive. In Apriori the decision makers provide preferences before the optimization process. The A posteriori approach search process determines a set of Pareto solutions. This set helps the decision makers to have a complete knowledge of the Pareto front. The Pareto front constitutes an acquired knowledge on the problem. Subsequently, the decision makers choose one solution from the set of solutions provided by the solver [9]. Interactive approach is a progressive interaction between the decision maker and the solver, by applying search components over the multiobjective metaheuristic methods.

A metaheuristic is formally outlined as an unvaried generation method that guides a subordinate heuristic by blending intelligence completely with distinct ideas for exploring and exploiting the search space, discovering strategies that are utilized to structure data in order to find efficiently near-optimal solutions [10]. Metaheuristic algorithms seek good solutions to optimization problems in circumstances where the complexity of the tackled problem or the search time available does not allow the use of exact optimization algorithms [11]. Solving MOPs with multiple global/ local optimal solutions by the evolutionary algorithms (EA) and metaheuristics deduce separation of a population of individuals into subpopulations, each connected to a different optimum, with the aim of maintaining diversity for a longer period.

Data analysis [12] plays an indispensable role for understanding various phenomena. Cluster analysis, primitive exploration with little or no prior knowledge, consists of research developed across a wide variety of mutliobjective metaheuristic methods [15]. Cluster analysis is all pervading in life sciences [13], while taxonomy is the term used to denote the activity of ordering and arranging the information in domains like botany, zoology, ecology, etc. The biological classification dating back to the 18th century (Carl Linne) and still valid, is just a result of cluster analysis. Clustering constitutes the steps that antecede classification. Classification aims at assigning new objects to a set of existing clusters/classes; from which point of view it is the 'maintenance' phase, aiming at updating a given partition. The goal of multiobjective clustering [14] is to find clusters in a data set by applying several Evolutionary-clustering algorithms corresponding to different objective functions. Optimization-based clustering algorithms [14] wholly rely on cluster validity indices, the optimum of which appears as a proxy for the unidentified "correct classification" during an antecedently unhandled dataset.

This survey provides an updated overview that is fully devoted to metaheuristic evolutionary algorithms for clustering, and is not limited to any particular kind of evolutionary approach, comprises of advanced topics in MOPs, MOEAs and data clustering. Section 1embarks the basic stipulations of MOPs and EC. Section 2 introduces relevant MOEAs classification scheme along with Pareto Optimality and briefly addresses the basic framework, algorithms, design, recent developments and applications in this field. Section 3 provides a catalog that foregrounds some significant vistas in the context of evolutionary data clustering. Section 4 summarizes the survey paper by addressing some important issues of MOPs, MOEAs and data clustering, and set most hopeful paths for future research.

# 2. MULTI OBJECTIVE EVOLUTIONARY ALGORITHM (MOEAs)

## 2.1. Introduction

MOEAs [7], [10], [14], [15] have involved a stack of erudition proposition during the last three decades, and they are still one of the most up-to-date study areas in the bifurcate of ECs. The first generation of MOEA [16] was characterized by rate of proposal entity based on Pareto ranking. The most frequent advancement to uphold diversity of the same is fitness sharing. In the other end, MOEAs are characterized by a stress on tidiness and by the site of selectiveness. Elitism [17] is a method to safeguard and use formerly found paramount results in succeeding generations of EA. Maintaining archives of non-dominated solutions is an important issue in elitism EA. The final contents of archive represent usually the result returned by optimization process.

When an area of concession (non-dominated) solutions is believed at search and managerial, these solutions involve a nutritious estimation to the Pareto optimal front [18]. The Pareto optimal front is the lay down of all non-subject solutions in the multiobjective coverage. However, when the solutions in the obtained specific do not huddle on the Pareto optimal front, then they must lessen to obtained non-dominated front or the informal Pareto front [6]. The pleasantness of Pareto optimization [14] emanates from the fact that, in most of the MOPs there is no such single best solution or global optima and it is very complicated to establish predilections among the criteria antecedent to the search. The spread and distribution of the non-dominated solutions, the proximity between the obtained front and the Pareto optimal front is decoratively obtained from the back issues of non-dominated solutions [10]. The three main goals set by Zitzler in [67], [71] could be achieved by MOPs are a) The Pareto solution set should be as close as possible to the true Pareto front, b) The Pareto solution set should be uniformly distributed and diverse over of the Pareto front, c) In order to provide the decision-maker a true picture of trade-offs, the set of solutions should capture the whole spectrum of the Pareto front. This needs inspection of solutions at the uppermost ends of the target function space.

**Definition1:** (MOP General Definition): In general, an MOP minimizes $F(\vec{x}) = (f_1(\vec{x}), \dots, f_k(\vec{x}))$ subject to $g_i(\vec{x}) \leq 0, i = 1, \dots, m, \vec{x} \in \Omega$. An MOP solution minimizes the component of a vector $F(\vec{x})$, where $\vec{x}$ is an n-dimensional decision variable vector ($\vec{x} = x_1, \dots, x_n$) from some universe $\Omega$. The MOPs evaluation function, $F: \Omega \rightarrow A$, maps decision variable ($\vec{x} = x_1, \dots, x_n$) to ($\vec{v} = a_1, \dots, a_k$). Using the MOP notation presented in Definition1, these key Pareto concepts are mathematically defined as follows:

**Definition 2:** (Pareto Dominance): A vector $\vec{u} = (u_1, \dots u_k)$ is said to dominate $\vec{v} = (v_1, \dots v_k)$ denoted by $\vec{u} \leq \vec{v}$ if and only if $u$ is partially less than $v$ i.e. $\forall i \in \{1, \dots, k\}, u_i \leq v_i \wedge \exists i \in \{1, \dots, k\}: u_i < v_i$.

**Definition 3:** (Pareto Optimality): A solution $x \in \Omega$ is said to be Pareto optimal with respect to $\Omega$ if and only if there us bi $x^{'} \in \Omega$ for which $\vec{v} = F(x^{'}) = (f_1(x^{'}), \dots, f_k(x^{'}))$ dominates $\vec{u} = F(x) = (f_1(x), \dots, f_k(x))$. The phrase "Pareto Optimal" is taken to mean with deference to the whole conclusion variable space except otherwise specified.

**Definition 4:** (Pareto Optimal Set): For a given MOP, $F(x)$, the Pareto Optimal Set $(\mathcal{P}^*)$ is defined as $\mathcal{P}^* \coloneqq \{x \in \Omega | \neg \exists \; x^{'} \in \Omega : F(x^{'}) \leq F(x)\}$.

**Definition 5:** (Pareto Front): For a given MOP, $F(x)$, the Pareto Optimal Set $(\mathcal{P}^*)$, the Pareto Front $(\mathcal{PF}^*)$ is defined as: $\mathcal{PF}^* \coloneqq \{\vec{u}\} = F(x) = (f_1(x), \dots, f_k(x)) | x \in \mathcal{P}^*$.

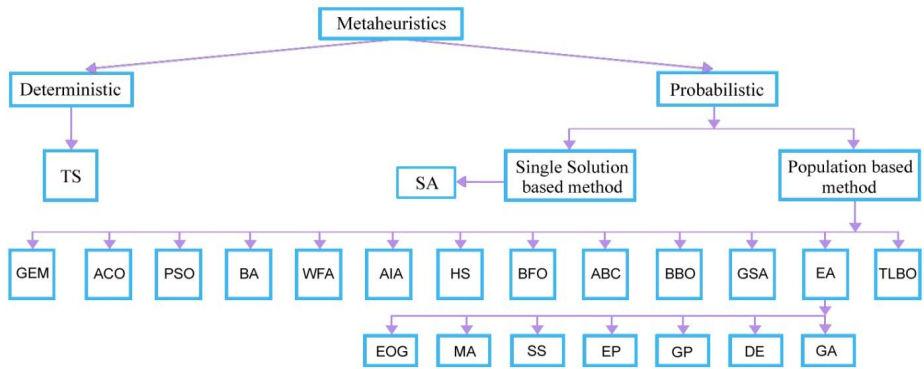

Figure 2. Taxonomy frameworks of meta-heuristics

While solving optimization problems, single-solution based metaheuristics improves a single solution in different domains. They could be viewed as walk through neighborhoods or search trajectories through the search space of the problem. The walks are performed by iterative dealings that move from the present answer to a different one within the search area. Population based metaheuristics [19] share the same concepts and viewed as an iterative improvement in a population of solutions. First, the population is initialized. Then, a new population of solutions is generated. It is followed with generation for a replacement population of solutions. Finally, this new population is built-in into present one using some selection procedures. The search method is stopped once a given condition is fulfilled. Figure 2 provides the taxonomy frameworks of multiobjective metaheuristics.

## 2.2. Deterministic meta-heuristics - Tabu Search (TS)

Tabu search [20], also known as Hill Climbing is essentially a sophisticated and improved type of local search, the algorithm works as follows: Consider a beginning accepted solution; appraise its adjoining solutions based on accustomed adjacency structure, and set the best as the first found neighbor, which is better than the present solution. Iterate the function until a superior solution is found in the neighborhood of the present solution. The bounded search stops if the present solution is better all its neighbors. The pseudo code 1 demonstrates the working procedure of Tabu Search.

Pseudo code 1: Tabu Search (TS)

---------------------------------------------------------------------------------------------

**Step 1:  Initialize**
**Step 2:  Repeat**
**Step 3:**          Generate all of the acceptable neighborhood solutions.
**Step 4**:          Evaluate the generated solutions.
**Step 5:**          Hold the acceptable ones as a candidate solution
**Step 6:**          If there is no suitable candidate then
**Step 7:**            Select the critical of the controlled solutions as a candidate
**Step 8:**          Update the tabu list.
**Step 9:**          Move to candidate solution.
**Step 10:**              If the number of engenerated solutions are sufficient, expand.
**Step 11: Until Termination** condition is met.

---------------------------------------------------------------------------------------------

## 2.3.Probabilistic meta-heuristics - Single solution based method

Simulated annealing (SA) is activated as the oldest element of meta-heuristic methods that gives single solution and corpuscles accumulation issues. Metropolis et al. proposed SA in 1953 [21]. It was motivated by assuming the concrete action of annealing solids. First, a solid is acrimonious to a top temperature and again cooled boring so that the arrangement at any time is about in thermodynamic equilibrium. At equilibrium, there may be abounding configurations with anniversary agnate to a specific activity level. The adventitious of accepting a change from the accepted agreement to a new agreement is accompanying to the aberration in activity amid with the two states. Since then, SA has been broadly acclimated in combinatorial optimization issues and it was proven that SA has accomplished acceptable after-effects on a variety of such issues. The pseudo code 2 demonstrates the working procedure of SA.

Pseudo code 2: Simulated Annealing (SA)

```
-----------------------------------------------------------
Step 1:  Initialize
Step 2:  Repeat
Step 3:          Generate a candidate solution.
Step 4:          Evaluate the candidate.
Step 5:          Determine the current solution.
Step 6:          Reduce the temperature.
Step 7: Until Termination condition is met.
-----------------------------------------------------------
```

## 2.4.Population based methods

Population based methods are those which not only impersonate the biological or natural occurrence but additionally establish with a set of initial realistic solutions called 'Population' and the intention is direct, that search in state space would to reach to the most favorable solution. EC are the recent search practices, which uses computational sculpt of procedure of evolution and selection. Perceptions and mechanism of Darwin [23] evolution and natural selection are predetermined in evolutionary algorithms (EAs) and are habituated to solve quandaries in many fields of engineering and science. EAs are recent, parallel, search and optimization practices, which uses computational sculpt of procedure of natural selection [23] and population genetics [24].

### 2.4.1. Evolutionary algorithms

EAs use the vocabulary borrowed from genetics. They simulate the evolution across a sequence of generations (iterations within an iterative process) of a population (set) of candidate solutions. A candidate solution is internally represented as a string of genes and is called chromosome or individual. The position of a gene in a chromosome is called locus and all the possible values for the gene form the set of alleles of the respective gene. The internal representation (encoding) of a candidate solution in an evolutionary algorithm forms the genotype that is processed by the evolutionary algorithm. Each chromosome corresponds to a candidate solution in the search space of the problem, which represents its phenotype. A decoding function is necessary to translate the genotype into phenotype. Mutation and Crossover are two frequently used operators referred to as evolutionary approaches [25].

Mutation consists in a random perturbation of a gene while crossover aims at exchanging genetic information among several chromosomes, thus avoids local optima. The chromosome subjected to a genetic operator is called parent and the resulted chromosome is called offspring. A process called selection involving some degree of randomness selects the individuals to Recombination or Crossover creates offspring's, mainly based on individual merit. The individual merit is evaluated using a fitness function, which quantifies how the candidate solution befitted being encoded by the chromosome, for the problem being solved. The fitness function is formulated based on the mathematical function to be optimized. The solution returned by an evolutionary algorithm is usually the most fitted chromosome in the last generation. The pseudo code 3 demonstrates the basic structure of Evolutionary Approaches and the general framework of Evolutionary Approaches is shown as Figure 3.

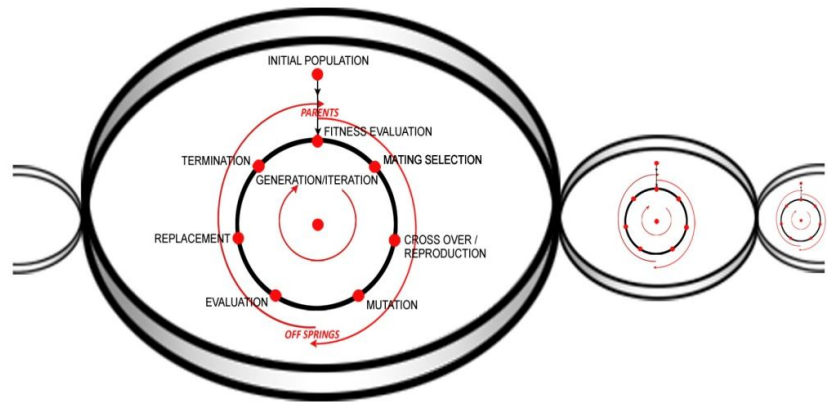

Figure 3. General Framework of Evolutionary Approaches

Several landmark methods identified in EAs are, Evolutionary Programming (EP) [35], [36], Genetic Programming (GP) [15], Differential Evolution (DE) [26], Scatter Search (SS) [27], Memetic Algorithm (MA) [28], and Genetic Algorithms (GA) [29], [30]. All these algorithms accept the genetic operations anchored central with accessory deviation. In recent times, heuristics or meta-heuristics are combined with these algorithms to form hybrid methods. Few independent implementation instances of EAs are GAs, developed by Holland [31] and Goldberg [32]. Evolution Strategies (ESs) developed by Rechenberg [33] and Schwefel [34], Evolutionary Programming (EP), originally developed by L. J. Fogel et al. [35] and subsequently refined by D. B. Fogel [36]. Each of these algorithms has been proved capable of yielding approximately optimal solutions when prearranged with complex, multimodal, non-differential, and discontinuous search spaces [37].

Pseudo code 3: Evolutionary Approaches (EA)
-----------------------------------------------------------------------------------------
**Step 1: Initialize**
**Step 2: Repeat**
**Step 3:** Evaluate the individuals.
**Step 4:** **Repeat**
**Step 5:** Select parents.
**Step 6:** Generate offspring.
**Step 7:** Mutate if enough solutions are generated.
**Step 8:** **Until** population number is reached.
**Step 9:** replicate the top fitted individuals into population as they were
**Step 10: Until** required number of generations are generate.
-----------------------------------------------------------------------------------------

### 2.4.2. Genetic algorithms (GAs)

Genetic algorithms (GA) [32], [38] are randomized search and optimization techniques guided by the attempts of change and accustomed genetics, and having a bulk of absolute parallelism. GAs perform search in intricate, immensely colossal and multimodal scenery, and provide near optimal solutions for fitness function of an optimization quandary. Evolutionary stress is applied with a stochastic method of roulette wheel in step 3; parent cull is utilized to pick parents for the incipient population. The canonical GA [39] encodes the problem within binary string individuals. The pseudo code 4 demonstrates the working procedure of GA. While selection is random any individual has the choice to become a parent, selection is clearly biased towards fitted individuals.

Parents are not required to be distinctive for any iteration; fit individuals may produce many offspring's the crossover is selected at random and mutation is applied to all individuals in the new population. With probability $P_m$, each bit on every string is inverted. The new population then becomes the current population and the cycle is repeated until some termination criteria satisfied [40]. The algorithm typically runs for some fixed number of iterations, or until convergence is detected within the population. The likelihood of mutation and crossover, $P_m$ and $P_c$ are factor of the algorithm and have to be set by the user [41].

Pseudo code 4: Genetic Algorithm (GA)

-----------------------------------------------------------------------------------------------------------

**Step 1:** A population of μ random individuals is initialized.
**Step 2:** Fitness scores are assigned to each individual.
**Step 3:** By means of roulette wheel select μ/2 pairs of parents from the existing
　　　　　　to form anatomy of new population.
**Step 4:** With probability $P_c$, offspring are produced by assuming crossover on the μ /2 pairs
　　　　　　of parents. The offspring alter the parents in the new population.
**Step 5:** With probability $P_m$, mutation is performed on the new population.
**Step 6:** The new population becomes the current population.
**Step 7:** If the extinction situation are annoyed exit, otherwise go to step 3.

-----------------------------------------------------------------------------------------------------------

### 2.4.3. Evolutionary strategies (ESs)

Historically ESs was designed for parameter optimization problems. The encoding used for each individual is a list of real numbers, called the object variables of the problem. The action ambit factors acclimates the behavior of the mutation operator and are necessary when translating to an individual [33], [34]. In each iteration, one offspring is generated from a population of size m. The same parents are used to acclimate to all object parameters in the child, and again the parents are re-selected for each action ambit factors. The parents are chosen randomly from the existing population. Mutation is the major abettor in ES and it acts upon action ambits as well as object parameters. Selection in EAs is deterministic i.e. the best m individuals are taken from the new offspring self-adaptation mechanism like (μ,λ) or (μ+λ) [55]. The pseudo code 5 demonstrates the working procedure of ES.

Pseudo code 5: Evolutionary Strategies (ESs)

-----------------------------------------------------------------------------------------------------------

**Step 1:** μ individuals are arbitrarily initialized from the existing population.
**Step 2:** All the μ individuals are dispensed with Fitness scores.
**Step 3:** λ New offspring are engendered by recombination from the current population.
**Step 4:** The λ new offspring are mutated.
**Step 5:** All the λ individuals are dispensed with Fitness scores
**Step 6:** A new population of m individuals is selected, using either (μ,λ) or (μ+λ) selection.
**Step 7:** The new population becomes the current population.
**Step 8:** If the termination conditions are convinced exit, else go to step 3.

-----------------------------------------------------------------------------------------------------------

### 2.4.4. Evolutionary programming (EP)

EP was first developed by L. J. Fogel et. al. [26] for the evolution of finite state machines using a restricted symbolic alphabet encoding. Individuals in EP comprise a string of real numbers. EP differs from GAs and ESs and there is no recombination operator. Evolution is wholly dependent on the mutation operator, using s a Gaussian probability distribution to perturb each variable. The self-adaption is used to control the Gaussian mutation operator. Tournament selection method is used to select fresh population from parents and children using mutation operator. The pseudo code 6 demonstrates the working procedure of EP.

Pseudo code 6: Evolutionary Programming (EP)

-------------------------------------------------------------------------------------------------------
**Step 1:** A current population of m individuals is randomly initialized.
**Step 2:** Fitness scores are assigned to each of the m individuals.
**Step 3:** m individuals in the existing population are used to create m children
using mutation operator
**Step 4:** Fitness scores are assigned to the m offspring.
**Step 5:** A new population of size m is created from the m parents and the m offspring
using Tournament selection.
**Step 6:** If the termination conditions are met exit, else go to step 3.
-------------------------------------------------------------------------------------------------------

### 2.4.5. Genetic programming (GP)

In GP individuals are computational programs. GP was developed by John Koza [40] and it is based on GA. GP is known as an successful research exemplar in Artificial Intelligence and Machine Learning, and have been studied in the most assorted areas of knowledge, such as data mining, optimization tasks etc. In GP, the EAs operate over a population of programs that have different forms and sizes. The primary population must have adequate multiplicity, that is, the individuals must have most of the distinctiveness that are essential to crack the crisis directed by a fitness function. Reproduction, Crossover, and Mutation genetic operators are applied after the selection to the each chromosome in population. Those chromosomes that are more preponderant solve the quandary, will receive a more preponderant fitness value, and accordingly will have a more preponderant chance to be culled for the next generation [42]. The objective of the GP algorithm is to cull, through recombination of genes, the program that that more preponderant solves a given quandary. The pseudo code 7 demonstrates the working procedure of GP.

Pseudo code 7: Genetic Programming (GP)

-------------------------------------------------------------------------------------------------------
**Step1:** Randomly create an initial population.
**Step 2: Repeat**
**Step 3:** Evaluate of each program by means of the fitness function.
**Step 4:** Select a subgroup of individuals to apply genetic operators.
**Step 5:** Apply the genetic operators.
**Step 6:** Replace the current population by this new population.
**Step 7: Until** a good solution or a stop criterion is reached.
-------------------------------------------------------------------------------------------------------

### 2.4.6. Ant colony optimization (ACO)

The first ACO [43] algorithm appeared in early 90s by was composed by Dorigo [44] is widely solving metaheuristic for combinatorial optimization problems. The abstraction is based on the ascertainment of foraging behavior of ants. When walking on routes from the nest to a source of food, ants seem to assume a acquisition of a simple desultory route, but a quite 'good' one, in agreement of shortness, or consistency, in terms of time of peregrinate. Thus, their deportment sanctions them to solve an optimization quandary [45]. The pseudo code 8 demonstrates the working procedure of ACO.

Pseudo code 8: Ant Colony Optimization (ACO)

```
-------------------------------------------------------------------------
Step 1:  Initialize
Step 2:  Set I to 0.
Step 3:  Repeat
Step 4:          Generate a feasible solution.
Step 5:          Evaluate goodness η.
Step 6:          If η > ηmax then
Step 7:                  Begin update Elite list.
Step 8:                      Shift the bounds End.
Step 7:  If I mod α = 0 then amend the result.
Step 8:  If I mod β = 0 then increase elite pheromone outlines.
Step 9:  Update pheromone trails.
Step 10: Increment I.
Step 11: Until i=σ.
-------------------------------------------------------------------------
```

Pseudo code 9: Particle Swarm Optimization (PSO)

```
-----------------------------------------------------------------------------
Step 1:  Initialize
Step 2:  Repeat
Step 3:          Evaluate fitness for each particle.
Step 4:          Update both the positions of local and global superlatives.
Step 5:          Update particle velocity by
                     v[i+1] = v[i]+c1*rand()*(pbest[i]-current[i])+c2*rand()*(gbest[i]-current[i]).
Step 6:          Update particle location by current[i+1] = current[i]+v[i].
Step 7:  Until maximum number of generation reached.
-----------------------------------------------------------------------------
```

### 2.4.7. Particle swarm optimization (PSO)

In mid-90s Kennedy and Eberhart [46], [47] first proposed the PSO algorithm. PSO is population-based and evolutionary in nature. This algorithm was stimulated by the allegory of communal interaction and correspondence in a flock of birds or school of fishes. In these communal groups, a central agent directs the association of the entire swarms. The association of each entity is based on the central agent and its own acquaintance. Hence, the particles in PSO algorithm rely on the central agent, which is assumed as the top performer [48], [49]. The pseudo code 9 demonstrates the working procedure of PSO.

### 2.4.8. Bees algorithm (BA)

D.T. pharm first proposed the Bee algorithm [50] as an optimization method stimulated by the usual foraging actions of honeybees to find the optimal result. The phenomenon behind this algorithm is the food foraging behavior of honeybees. Honeybees are usually able to elongate their colony over stretched spaces and in sundry probable directions, concurrently to capitalize on significant number of food sources. A colony is prospered by reallocating its foragers to their opportune fields. Typically, more bees are recruited for flower patches with ample amounts of nectar or pollen that can be accumulated with less exertion [51]. The pseudo code 10 demonstrates the working procedure of BA.

Pseudo code 10: Bees Algorithm (BA)

-------------------------------------------------------------------------------------------------
**Step 1: Initialize**
**Step 2: Repeat**
**Step 3:**           Evaluate fitness of the population.
**Step 4:**           **While** (stopping criterion did not meet)
**Step 5:**              Select sites for neighborhood search.
**Step 6:**              Recruit bees for selected sites (more bees for the best sites).
**Step 7:**              Evaluate fitness.
**Step 8:**              Select the fittest bee from each site.
**Step 9:**              Assign remaining bees to search randomly and evaluate their fitness.
**Step 10:**          **End While**
**Step 11: Until** maximum number of generation reached.
-------------------------------------------------------------------------------------------------

Pseudo code 11: Water Flow-like Algorithm (WFA)

---------------------------------------------------------------------------------------------
**Step 1: Initialize**
**Step 2: Repeat**
**Step 3:**           **Repeat**
**Step 4:**              Calculate number of sub flows.
**Step 5:**              For each sub flow find best neighborhood solution.
**Step 6:**              Distribute mass of flow to its sub flows.
**Step 7:**              Calculate improvement in objective value.
**Step 8:**           **Until**   Population no. reached.
**Step 9:** Merge sub flows with same objective values.
**Step 10:** Update the no. of sub flows.
**Step 11:** Update total no. of water flows.
**Step 12: If** Precipitation condition met
**Step 13:**              Perform bit reordering strategy.
**Step 14:**              Distribute mass to flows.
**Step 15:** Evaluate new solution.
**Step 16:** Update the no. of sub flows.
**Step 17:** Update total no. of water flows.
**Step 18: Until** maximum number of generation reached.
---------------------------------------------------------------------------------------------

### 2.4.9. Water flow-like algorithm (WFA)

WFA was proposed by Yang [52] as a nature inspired optimization algorithm for object clustering. It impersonate the action of water flowing from higher to lower level and aids in the process of searching for optimal result. The pseudo code 11 demonstrates the working procedure of WFA.

### 2.4.10. Differential evolution

DE [29], [53] materialized as a straightforward and well-organized scheme for global optimization over continuous spaces more than a decade ago. It was kindred by computational procedures as engaged by the typical evolutionary application to multiobjective, constrained, large scale, and uncertain optimization issues [45]. DE solves the objective function by twisting and tuning the sundry with the ingredient like initialization, mutation, diversity enhancement, and selection as well as by the cull of the control parameters etc. [54]. The pseudo code 12 demonstrates the working procedure of DE.

Pseudo code 12: Differential Evaluation (DE)

---------------------------------------------------------------------------------------------------------
**Step 1:  Begin**
**Step 2:**  Generate randomly an initial population of solutions.
**Step 3:**  Calculate the fitness of the initial population.
**Step 4:  Repeat**
**Step 5:**      Arbitrarily choose three solutions for each parent
**Step 6:**      Create one offspring using the DE operators.
**Step 7:**      repeat the procedure until it is equal to population size
**Step 8:**      For each member of the next generation.
**Step 9:**      If children (x) is more shaped than parent (x).
**Step 10:**      Parent (x) is replaced.
**Step 11: Until** a stop condition is satisfied.
**Step 12: End.**
---------------------------------------------------------------------------------------------------------

Some of the well-known meta-heuristics developed during the last three decades are Artificial Immune Algorithm (AIA) [55], which works on the immune system of the human being. Music creativeness in a music player is described as Harmony Search (HS) [56], the deeds of bacteria is shown as Bacteria Foraging Optimization (BFO) [57]. The messaging between the frogs is illustrated in Shuffled Frog Leaping (SFL) [58]. The algorithm for species movements like migration and colonization is shown in Biogeography Based Optimization (BBO) [59]. The theory of gravitational force, staging between the groups of bodies is described in Gravitational Search Algorithm (GSA) [60]. The law of detonation of a grenade is shown in Grenade Explosion Method (GEM) [61]. These algorithms are applied to numerous engineering optimization issues and proved effective in solving precise kind of problems.

Teaching Learning Based Optimization (TLBO) [15] is an optimized method used to acquire global solutions for continuous non-linear functions with fewer computational exertion and high reliability. The viewpoint between teaching and learning is well exhibited in TLBO algorithm. The TLBO algorithm is based on the influence of a teacher on the output of learners in a class. The superiority of a teacher influences the outcome of learners. It is evident that a good teacher trains learners such that they can have better outcome in terms of their marks or grades. Besides, learners also learn from communication amongst themselves, which also facilitate in improving their results. All EAs carve up the same basic idea, but diverge in the way they encode the result and on the operators they use to produce the next generation. EAs are restricted by numerous inputs, such as the size of the population, the rates that control how often mutation and crossover are used. Broadly, there is no assurance that the EAs will find the optimal result to any subjective problems, but careful handling of inputs and choosing an appropriate algorithm is passable to the problem to increase the chances of accomplishment [58].

MOEA based on decomposition (MOEA/D) [17] is a recent framework based on traditional aggregation methods that decomposes the MOP into a number of Scalar Objective Optimization problems (SOPs). The aim of each SOP, or a sub problem, is a (linearly or nonlinearly) weighted aggregation of the individual objectives. Memetic MOEA methods adept to offer not only more preponderant speed of convergence to EAs, but with more preponderant correctness for the results [62]. Ishibuchi and Murata suggested one of the first memetic MOEAs [63]. The algorithm utilizes a local search procedure after the traditional variation operators are directed, and arbitrarily they drawn scalar function, to accredit fitness, for parent selection.

MOPs are used to embark upon the arduousness of fuzzy partitioning where a number of fuzzy cluster validity indices are simultaneously optimized. The resultant set of near-Pareto-optimal answers contains a number of non-dominated answers, which the user can referee relatively and choose upon the most undertaking one according to the problem necessity; Real-coded encoding of the cluster centers is utilized for this principle [64].

The well-relished advances that have been utilized by distinct researchers for MOPs are: Weighted-sum –Approach [67], Schaffer's Vector Evaluated Generic Algorithm (VEGA) [48], Fonseca and Fleming's Multi-Objective Genetic Algorithm (MOGA) [49], Horn, Nafploitis and Goldberg's Niched Pareto Genetic Algorithm (NPGA) [65], Zitzler and Thiele's strength Pareto Evolutionary Algorithm (SPEA) [66], Srinivas and Deb's Non-dominated Sorting Genetic Algorithm (NSGA) [68] is used to determine the appropriate cluster centers and the corresponding partition matrix [69], NSGA-II [101], Vector Optimized Evolution Strategy (VOES) [71] , Weight-based Genetic Algorithm (WBGA) [72], Predator-Prey evolution strategy (PPES) [73], Rudolph's Elitist multiobjective  Evolutionary Algorithm [75], Knowles and Corne suggested Pareto-archived evolution strategy (PAES) [54], which uses an evolutionary strategy.

Multi-objective genetic local search algorithm (MOGLS) for solving combinatorial multiobjective optimization problems was first proposed by Ishibuchi and Murata [97]. MOGLS applies a local search technique after using standard variation operators. A scalar fitness function was utilized to cull a couple of parent solutions to engender incipient solutions with crossover and mutation operator. Hybrid evolutionary metaheuristics (HEMH) [77] presents a combination with different metaheuristics, each one integrated to other to improve the search capabilities. The hybridization of greedy randomized adaptive search procedure (GRASP) with data mining (DM-GRASP) [78] is directed to a primary set with high quality solutions discrete along the Pareto front within the framework of MOEA/D.

To handle dynamic MOPs, a new co-evolutionary algorithm (COEA) is suggested [79]. It hybridizes competitive and supportive procedures experiential in world to pathway the Pareto front in a dynamic environment. The main conception of the competitive-cooperative co-evolution is to permit the decomposition process of the optimization quandary to habituate and emerge rather than being hand-designed and fine-tuned at the commencement of the evolutionary optimization procedures. Non-dominated arranging Genetic Algorithm-II (NSGA-II) [80] is utilized to work out the befitting cluster centers and the corresponding partition matrix.

### 2.4.11. Applications

As MOEAs recognition is rapidly grown, as successful and robust multiobjective optimizers and researchers from various fields of science and engineering have been applying MOEAs to solve optimization issues arising in their own domains. The domains where the MOEAs optimization methods [81] applied are Scheduling Heuristics, Data Mining, Assignment and Management, Networking, Circuits and communications, Bioinformatics, Control systems and robotics, Pattern recognition and Image Processing, Artificial Neural Networks and Fuzzy systems, Manufacturing, Composite Component design, Traffic engineering and transportation, Life sciences, embedded systems, Routing protocols, algorithm design, Web-site and Financial optimization. Few Prons and Corns that are faced by MOEAs while solving the optimization methods are:

a.  The problem has multiple, possibly incommensurable, objectives.
b.  The Computational time for each evaluation is in minutes or hours
c.  Parallelism is not encouraged

d.   The total number of evaluations is limited by financial, time, or resource constraints.
e.   Noise is low since repeated evaluations yield very similar results
f.   The overall decline in cost accomplished is high.

## 3.  DATA CLUSTERING

Data is usually in the raw form, Information is processed into data i.e. a semantic connection gives the data a meaning.  Records called data items are expressed as tuples of numerical/categorical values; each value in the tuple indicates the observed value of a feature. The features in a data set are also called attributes or variables. Information can be automatically extracted by searching patterns in the data. The process of detecting patterns in data is called learning from data. Extorting implicit, previously unknown and impending useful information from data comprise the substance of the Data Mining.  The algorithmic framework providing automatic support for data mining is generally called Machine Learning [83].

Association rule mining aims at detecting any association among the features. Classification aims at predicting the value of a nominal feature called the class variable. Classification is called supervised learning because the learning scheme is presented with a set of classified examples from which it is expected to learn a way of classifying unseen examples. Data Clustering is executed in two different modes called crisp and fuzzy. Classification is overseen as a supervised learning because the learning scheme is presented with a set of classified examples from which it is anticipated to discover a way of classifying unseen examples. In crisp clustering, the clusters are disjoint and non-overlapping; any sample may fit in to one and only one category. In fuzzy clustering, a sample may belong to more than one category with a certain fuzzy membership ranking [84].

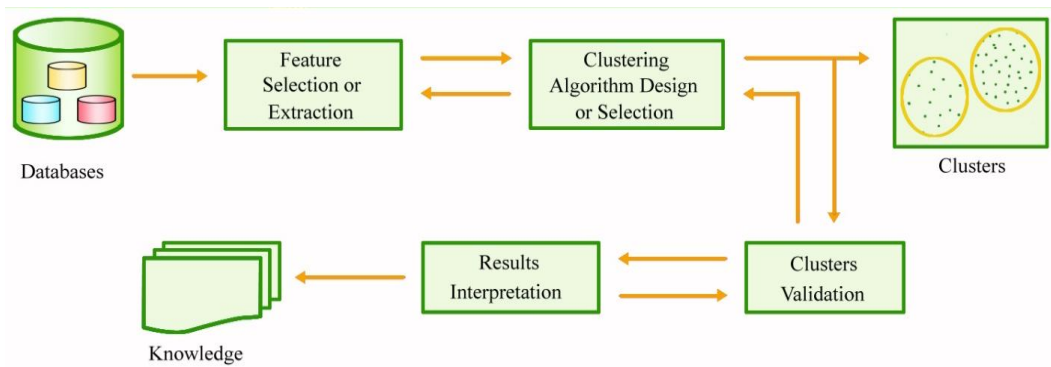

Figure 4.  Delineate of Clustering

Cluster analysis includes a series of steps, ranging from preprocessing i.e. feature selection, algorithm development, cluster validity, evaluation and interpretation. Each of them is closely associated to each other and wield huge confront to the methodical disciplines in mining knowledge from datasets [85], [86]. The clustering delineate is exemplified in Figure 4.  At both preprocessing and post-processing stage, feature selection/extraction and cluster validation are important in any clustering algorithms. Preferring suitable and significant features can significantly lessen the difficulty of consequent blueprint and result assessments to reveal the degree of assurance to which it can rely on the generated clusters. However, both processes lack universal guidance. Eventually, the substitution of different evaluations and technique is still reliant on the applications themselves [85].

Clustering is ubiquitous, and wealth of clustering algorithms that has been developed to solve different problems in specific fields. There are no intensely wide-ranging encroachments in clustering that can be globally used to solve all problems [87]. In general, algorithms are premeditated with certain supposition and favor some type of prejudice. Selection of initial seeds greatly affects the quality of the clusters and in k-means type algorithms. Most of the seed selection methods consequence different results in different independent runs. It was proven in [88] by K Karteeka Pavan et.al. a single optimal, outlier insensitive seed selection algorithm for k-means algorithms (SPSS) as extension to k-means++, resulted effectiveness in clustering when demonstrated on synthetic, real and on microarray data sets.

Clustering is fully functional in an extensive diversity of fields, ranging from engineering, machine learning, artificial intelligence, pattern recognition, computer sciences, life and medical sciences, earth sciences, social sciences, etc. [90]. The sketch out of unsupervised predictive learning is as follows: Distance and Similarity Measures [68], Hierarchical [69] (Agglomerative-BIRCH, CURE, ROCK, Divisive - DIANA, MONA), Squared Error [91] (Vector Quantization, K-Means, GKA,PAM), PDF Estimation via Mixture Densities [92] (GMDD), Graph Theory [93] (Chameleon, DTG, HCS, CLICK, CAST), Combinatorial Search Techniques [94] (GGA, TS and SA clustering), Fuzzy [95] (FCM, MM, PCM, FCS), Neural Networks [96] [97] (LVQ, SOFM, ART, SART, HEC, SPLL), Kernel [98] (SVC), Sequential Data [99] (sequence clustering, Statistical sequence clustering), Large-Scale Data [100] (CLARA, CURE, CLARANS, BIRCH, DBSCAN, DENCLUE, Wave Cluster, FC, ART), Data visualization and High-dimensional Data [101] (PCA, ICA, LLE, CLIQUE, OptiGrid, ORCLUS).

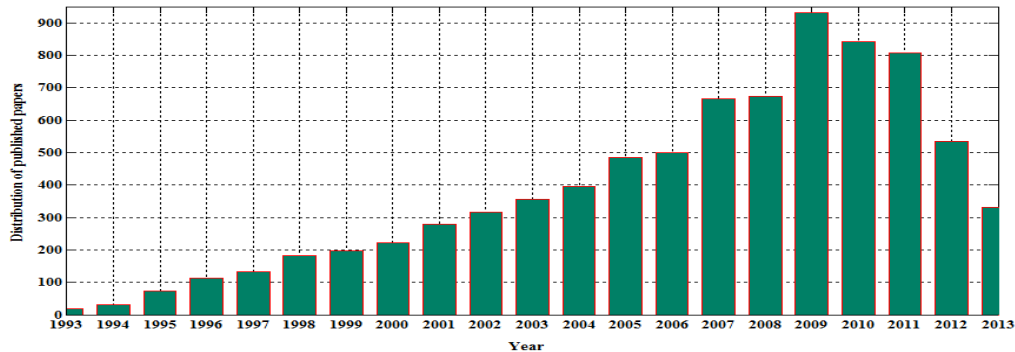

Figure 5. Year- wise distribution of MOEA publications

MOEA predicated soft subspace clustering (MOSSC) is directed to popular elitist multiobjective genetic algorithm NSGA-II. MOSSC [14] can concurrently optimize the weighting within-cluster compression and weighting between-cluster partition integrated within two different clustering validity criteria. Co-clustering [32] endeavor at clustering both the samples and the characteristics concurrently to recognize hidden impede organization implanted into the data matrix. Clustering ensembles have materialized as a high-flying procedure for advancing robustness, steadiness and accuracy of unsupervised classification solutions [102]. The concurrency of clustering algorithms is foreordained, implemented and applied to numerous applications. Majority of the clustering task needs iterative procedures to find locally or globally optimal solutions from high-dimensional data sets.

The research on MOEA has its origin in 1967, but a remarkable amount of publications were not endorsed until 1990's, from then onwards, the investigation on MOEA equipped drastically. Figure 5 and Figure 6 attempts to envision the overall journeying on MOEA that is listed in EMOO statistical Repository [103] and maintained by Professor CA Coello. By March 2013, more than 8109 publications have been published in MOEA including data clustering [103].

Fig.6 envisages MOEAs publications into categories, the distribution includes 44% journal papers, 39% conference papers, 10% in-collection literature, 4% published Ph.D Thesis, and 1% of Master Thesis, Technical Reports and Books on MOEA. Fig. 5 envisages year-wise distribution of MOEAs publications in all categories. The first cohort of MOEA flanked by the years 1993-1999 was exemplifying by the minimalism of the algorithm. The second cohort sandwiched between the years 2000-2009 concerned about MOEAs competence both at algorithmic and data structure level. The third cohort is on track since the year 2010, addressing the bio inspired metaheuristic algorithms, elitism, Pareto ranking schemes, etc. The average numbers of publications that have been published in first, second and third cohort were 107, 483, 629 respectively. Hence, it is pragmatically an optimistic inclination that MOEA is still in premature arena and gigantic volumes of novel research algorithms in MOEA and data clustering are yet to be published.

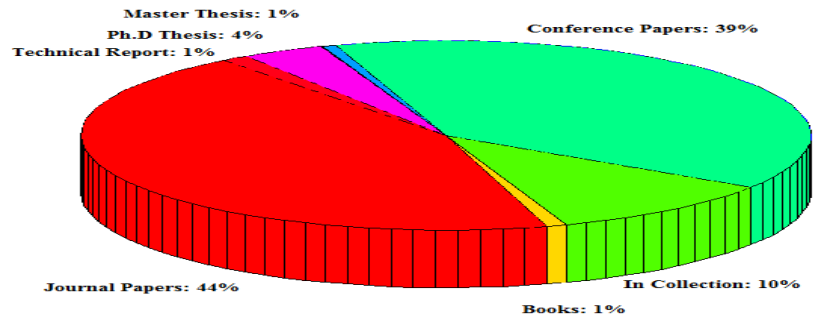

Figure 6. Distribution of MOEA publications by categories

This study focused on clustering algorithms developed with multiobjective metaheuristic methods embedded with the flavor of evolutionary algorithms. The validity indices determine the optimal number of clusters and corresponds the best cluster structure, whether the numbers of their cluster centers are same or not [89]. Numerous sprints with distinct initializations or factors capitulate multiple solutions. The paramount of these solutions is often culled by cluster validity measures.

## 4. CONCLUSION AND FUTURE RESEARCH DIRECTIONS

The present survey endeavors to provide a wide-ranging impression of work that has been done in the last three decades in MOPs, MOEAs in line with data clustering. The survey includes basic historical framework of MOEAs, algorithms, methods to maintain diversity, advances in MOEA designs, MOEAs for complicated MOPs, benchmark problems, methodological issues, applications and research inclinations.

An imperative surveillance was success of most MOEAs depends on the careful balance of two conflicting goals, exploration (searching new Pareto Optimal Solution) and exploitation (refining the obtained Pareto Solutions). EAs are easy to portray and execute, but rigid to analyze hypothetically. In spite of much experiential acquaintance and successful application, only little theoretical fallout pertaining to their effectiveness and competence are available.

Optimization based data clustering methods wholly rely on a cluster validity indices, the best of which appears as a substitute for the unidentified "correct classification" in a previously unhandled dataset. Incipient technology has engendered more intricate and exigent tasks, requiring more potent powerful clustering algorithms.

Evolutionary MOPs are still in early stages, and several research issues that remain yet to be solved are:

a. Rationalizing of MOEAs convention in real world problems
b. The stopping conditions of MOEAs, since it is not conspicuous to estimate when the population has reached a point from which no additional enhancement can be accomplish.
c. Identifying the paramount solution from Pareto Optimal Set
d. Hybridization of MOEAs
e. The influence of mating restrictions on MOEAs
f. Effectiveness and robustness in searching for a set of global trade-off solutions.

Although a lot of research work has been done in multibojective metaheuristics optimization in the recent years, still the theoretical fragment of MOEAs encroachment is not so much browbeaten. Inconceivable examination on different fitness assignment methods, validities, pattern distinctiveness, performance metrics blending with different selection schemes of EAs and data clustering are yet to be investigated.

# REFERENCES

[1] A. Mackworth, R. Goebel, Computational Intelligence: A Logical Approach by David. Oxford University Press. ISBN 0-19-510270-3.
[2] R. Kicinger, T. Arciszewski, K.A. De Jong, (2005) "Evolutionary computation and structural design: a survey of the state of the art", Computers & Structures, Vol. 23-24, pp. 1943-1978.
[3] H.P. Schwefel, (1977) "Numerische Optimierung von Computer-modellen mittels der Evolutionsstrategie", Basel: Birkhaeuser Verlag.
[4] D.E. Goldberg, (1989) Genetic Algorithms in Search, Optimization and Machine Learning, Addison-Wesley Publishing Company, Reading, Massachusetts.
[5] H. C. Martin Law, P. Alexander, K. A. Jain, (2004) "Multiobjective Data Clustering", IEEE Computer Society Conference on Computer Vision and Pattern Recognition.
[6] R.T. Marler, J.S. Arora, (2004) "Survey of multi-objective optimization methods for engineering", structured multidisc optimum, Vol. 26, pp. 369–395, DOI 10.1007/s00158-003-0368-6.
[7] V. Pareto, (1906) "Manuale di Economia Politica. Societ`a Editrice Libraria", Milan.
[8] A. K. Jain, M. N. Murty, P. J. Flynn, (1999) "Data clustering: a review". ACM Computer. Survey, Vol. 31, No. 3, pp.264–323.
[9] Z. Ya-Ping, S. Ji-Zhao, Z. Yi,Z. Xu, (2004) "Parallel implementation of CLARANS using PVM. Machine Learning and Cybernetics", Proceedings of 2004 International Conference on Cybernetics, Vol. 3, pp. 1646–1649.
[10] I.H. Osman, G. Laporte, ( 1996) "Metaheuristics: a bibliography", Annals of Operations Research, Vol. 6, pp. 513–623.
[11] R.V. Rao, V.J. Savsani, D.P. Vakharia, (2012) "Teaching–learning-based optimization: an optimization method for continuous non-linear large scale problems", Information Sciences, Vol. 183, pp. 1–15.
[12] X. Rui, D. Wunsch II, (2005) " Survey of Clustering Algorithms", IEEE Transactions On Neural Networks, Vol. 16, No. 3.
[13] A. Fred, A. K. Jain, (2002) "Evidence accumulation clustering based on the k-means algorithm." Structural, syntactic, and statistical pattern recognition. Springer Berlin Heidelberg, Vol. 23, pp. 442-451.
[14] C.A. Coello Coello, (1996) "An Empirical Study of Evolutionary Techniques for Multiobjective Optimization in Engineering Design". Ph.D. Thesis, Department of Computer Science, Tulane University, New Orleans, LA.
[15] C. Dimopoulos, (2006) "Multi-objective optimization of manufacturing cell design". International Journal of Production Research, Vol. 44, No. 22, pp. 4855-4875.

[16] K. Deb, (2001) Multi-objective Optimization Using Evolutionary Algorithms, John Wiley and Sons Ltd, England.

[17] Q. Zhang, H. Li, (1997) "MOEA/D: a Multiobjective evolutionary algorithm based on decomposition", IEEE Transactions on Evolutionary Computation, Vol. 11, No. 6, pp. 712–73.

[18] R. Kumar, P. Rockett, (2002) "Improved Sampling of the Pareto-Front in Multiobjective Genetic Optimizations by Steady-State Evolution: A Pareto Converging Genetic Algorithm", Evolutionary Computation, Vol. 10, No. 3, pp. 283-314.

[19] T. Ghosha, S. Senguptaa, M. Chattopadhyay, P. K. Dana, (2011) "Meta-heuristics in cellular manufacturing: A state-of-the-art review", International Journal of Industrial Engineering Computations, Vol. 2, pp. 87–122.

[20] F. Glover, M. Laguna, "Tabu Search", Kluwer Academic Publishers, Norwell, MA, USA, 1997.

[21] Metropolis, A. Rosenbluth, M. Rosenbluth, A. Teller, E. Teller, (1953) "Equation of State Calculations by Fast Computing Machines", J. Chem. Phys. Vol. 21, No. 6, pp. 1087-1092.

[22] Kirkpatrick, E. Aarts, J. Korst, "Simulated Annealing and the Boltzmann Machine". John Wiley & Sons, New York, USA.

[23] C. Darwin, (1929) The origin of species by means of Natural selection or the preservation of avowed races in the struggle for life, The book League of America, Originally published in 1859.

[24] R.A. Fisher, (1930) The genetical theory of natural selection, Oxford: Clarendon Press.

[25] T. Back, D. B. Fogel, Z. Michalewicz, (1997) Handbook of Evolutionary Computation, U.K., Oxford University Press.

[26] L. J. Fogel, A. J. Owens, M. J. Walsh, (1966) Artificial Intelligence through Simulated Evolution, Wiley, New York.

[27] D. Karaboga, (2005) "An Idea Based on Honey Bee Swarm for Numerical Optimization, Technical Report-TR06", Erciyes University, Engineering Faculty, Computer Engineering Department.

[28] A. Muruganandam, G. Prabhaharan, P. Asokan, V. Baskaran, (2005) "A memetic algorithm approach to the cell formation problem", International Journal of Advanced Manufacturing Technology, Vol. 25, pp. 988–997.

[29] M.G. Epitropakis, (2011) "Proximity-Based Mutation Operators", IEEE Transactions on Evolutionary Computation, Vol. 15, No. 1, pp 99-119.

[30] D.E. Goldberg, (1989) Genetic Algorithms in Search Optimization and Machine Learning. Addison Wesley.

[31] J. H. Holland, (1982) Adaptation in Natural and Artificial Systems, MIT Press, Cambridge, MA.

[32] D. E. Goldberg, K. Swamy, (2010) Genetic Algorithms: The Design of Innovation, 2nd, Springer, 2010.

[33] I. Rechenberg, (1973) Evolutionsstrategie: Optimierung Technischer Systeme nach Prinzipien der Biologischen Evolution, Frommann- Holzboog, Stuttgart.

[34] H.P. Schwefel, (1981) Numerical Optimization of Computer Models, Wiley, Chichester.

[35] L.J. Fogel, G.H. Burgin, (1969) "Competitive goal-seeking through evolutionary programming," Air Force Cambridge Research Laboratories.

[36] D. B. Fogel, (1995) Evolutionary Computation: Toward a New Philosophy of Machine Intelligence, IEEE Press, Piscataway, NJ.

[37] G. Jones, Genetic and Evolutionary Algorithm, Technical Report, University of Sheffield, UK.

[38] D. E. Goldberg, K. Deb (1991) "A comparative analysis of selection schemes used in genetic algorithms." Urbana 51.

[39] J. Holland, (1975) Adaptation in Natural and Artificial Systems, University of Michigan Press, Ann Arbor.

[40] J. Koza, (1992) Genetic programming: on the programming of computers by means of natural selection, MIT Press, Cambridge.

[41] J. Liu, L. Tang, (1999)"A modified genetic algorithm for single machine scheduling", Computers & Industrial Engineering,Vol. 37, pp. 43–46.

[42] M. C. da Rosa Joel, (2009) "Genetic Programming and Boosting Technique to Improve Time Series Forecasting, Evolutionary Computation" Wellington Pinheiro dos Santos, ISBN: 978-953-307-008-7.

[43] M. Dorigo, (1992) Optimization, Learning and Natural Algorithm", Ph.D. Thesis, Politecnico di Milano, Italy.

[44] M. Dorigo, T. Stutzle, (2004) Ant Colony Optimization. MIT Press, Cambridge, MA, USA.

[45] S Das, N. S. Ponnuthurai, (2011) "Differential Evolution: A Survey of the State-of-the-Art", IEEE Transactions On Evolutionary Computation, Vol. 15, No. 1.

[46] C.L. Sun, J.C. Zeng, J.S. Pan, (2011) "An improved vector particle swarm optimization for constrained optimization problems", Information Scieces, Vol. 181, pp. 1153–1163.

[47] J. Kennedy, R.C. Eberhart, (1995) "Particle swarm optimization", in Proceedings of IEEE international conference on neural networks, pp. 1942–194.

[48] J. David Schaffer, (1995) "Multiple Objective Optimization with Vector Evaluated Genetic Algorithms", Proceedings of the 1st International Conference on Genetic Algorithm, pp. 93-100, ISBN:0-8058-0426-9.

[49] C. M. Fonseca, P. J. Fleming, (1993) "Genetic algorithms for multiobjective optimization: Formulation, discussion and generalization", in Proceedings of the Fifth International Conference on Genetic Algorithms, S. Forrest, Ed. San Mateo, CA: Morgan Kauffman, pp. 416–423.

[50] D.T. Pham et.al. (2006) "The bees algorithm – A novel tool for complex optimization problems", Proceedings of IPROMS Conference, pp. 454– 461.

[51] A. Ahrari, A.A. Atai, (2010) "Grenade explosion method – a novel tool for optimization of multimodal functions", Applied Soft Computing, Vol. 10, pp. 132–1140.

[52] F.C. Yang, Y.P. Wang, (2007) "Water flow-like algorithm for object grouping problems", Journal of the Chinese Institute of Industrial Engineers, Vol. 24, No. 6, pp. 475–488.

[53] R. Storn, K. Price,(2011) "Differential evolution – a simple and efficient heuristic for global optimization over continuous spaces", Journal of Global Optimization, Vol. 11, pp. 341–359.

[54] J.D. Knowles, D.W. Corne, (2000) "Approximating the non-dominated front using the Pareto archived evolution strategy", Evolutionary Computing, Vol. 8, pp. 142-172.

[55] J.D. Farmer, N. Packard, A. Perelson, (1989) "The immune system, adaptation and machine learning", Physica D, Vol. 22, pp. 187–204.

[56] A.R. Yildiz, (2009) "A novel hybrid immune algorithm for global optimization in design and manufacturing", Robotics and Computer-Integrated Manufacturing, Vol. 25, pp. 261–270.

[57] K.M. Passino, (2002) "Biomimicry of bacterial foraging for distributed optimization and control", IEEE Control Systems Magazine, Vol. 22, pp. 52–67.

[58] M. Eusuff, E. Lansey, (2003) "Optimization of water distribution network design using the shuffled frog leaping algorithm", Journal of Water Resources Planning and Management ASCE, Vol. 129, pp. 210–225.

[59] D. Simon, (2008) "Biogeography-based optimization", IEEE Transactions on Evolutionary Computation, Vol. 12, 702–713.

[60] E. Rashedi, H.N. Pour, S. Saryazdi, (2009) "GSA: a gravitational search algorithm", Information Sciences, Vol. 179, 2232–2248.

[61] J. Gareth, Genetic and Evolutionary Algorithms, Technical Report, University of Sheffield, UK.

[62] A. Lara, G. Sanchez, C.A. Coello Coello, O. Schutze, (2010) "HCS: a new local search strategy for memetic multiobjective evolutionary algorithms", IEEE Transactions on Evolutionary Computation, Vol. 14, No. 1, pp. 112–132.

[63] H. Ishibuchi, T. Murata, (1998) "Multi-Objective Genetic Local Search Algorithm and Its Application to Flowshop Scheduling", IEEE Transactions on Systems, Man and Cybernetics, Vol. 28, No. 3, 392–403.

[64] S. Bandyopadhyay, U. Maulik, A. Mukhopadhyay, "Multiobjective Genetic Clustering for Pixel Classification in Remote Sensing Imagery", IEEE Transactions on Evolutionary Computation.

[65] A. Bill, Xi. Wang, M. Schroeder, (2009) "A roadmap of clustering algorithms: finding a match for a biomedical application". Brief Bioinform.

[66] J. Horn, N. Nafploitis, D. E. Goldberg, (1994) "A niched Pareto genetic algorithm for multiobjective optimization" in Proceedings of the First IEEE Conference on Evolutionary Computation, Z. Michalewicz, Ed. Piscataway, NJ: IEEE Press, pp. 82–87.

[67] E. Zitzler, L. Thiele, (1998) "Multiobjective optimization using evolutionary algorithms—A comparative case study" in Parallel Problem Solving From Nature, Germany: Springer-Verlag, pp. 292–301.

[68] B. Scholkopf, A. Smola, (2002) "Learning with Kernels: Support Vector Machines, Regularization, Optimization, and Beyond", Cambridge, MA: MIT Press.

[69] N. Srinivas, K. Deb, (1995) "Multiobjective function optimization using nondominated sorting genetic algorithms," Evolutinary Computing, Vol. 2, No.3, pp. 221–248.

[70] J. T. Tou, R. C. Gonzalez, (1974) Pattern Recognition Principles Addison-Wesley.

[71] E. Zitzler, K. Deb, L. Thiele, (2000) "Comparison of multiobjective evolutionary algorithms: Empirical results", Evolutionary Computing, Vol. 8, pp. 173-195.

[72] F.A. Kursawe, (1990) "Variant of evolution strategies for vector optimization", in Parallel Problem Solving from Nature", H.-P. Schwefel and R. Manner, Eds., Berlin, Germany: Springer, pp. 193-197.

[73] P. Hajela, J. Lee, "Constrained genetic search via search adaptation. An Immune network solution", Structural optimization, Vol. 13, pp. 111-115.

[74] M. Luamanns, G. Rudolph, (1998) "A Spatial Predator-Prey approach to Multiobjective optimization a preliminary Study", in Proceedings on the Parallel Problem solutions from nature, pp. 241-249.

[75] G. Rudolph, (1998) "Evolutionary search for minimal elements on partially ordered finite sets", Evolutionary Programming, Vol. 8, pp. 345-353.

[76] H. Ishibuchi, T. Murata, (1998) "A multiobjective genetic local search algorithm and its application to flow shop scheduling", IEEE Transactions on Systems, Man, and Cybernetics Part C: Applications and Reviews, Vol. 28 No. 3, pp. 392–403.

[77] A. Kafafy, A. Bounekkar, S. Bonnevay, (2011) "A Hybrid Evolutionary Metaheuristics (HEMH) applied on 0/1 Multiobjective Knapsack Problems", in Proceedings of the 13th annual conference on Genetic and evolutionary computation, GECCO '11, ACM, New York, NY, USA, pp. 497–504.

[78] M. H. Ribeiro, A. Plastino, S. L. Martins, (2006) "Hybridization of GRASP Metaheuristic with Data Mining Techniques", J. Math. Model. Algorithms, Vol. 5, No. 1, pp. 23–41.

[79]C. K. Goh, K. C. Tan, (2009) "A Competitive-Cooperative Coevolutionary Paradigm for Dynamic Multiobjective Optimization", Trans. Evol. Comp, Vol. 13, pp. 103–127.

[80] M. H. Ribeiro, A. Plastino, S. L. Martins, (2005) "A Hybrid GRASP with Data Mining for the Maximum Diversity Problem", Second International Workshop on Hybrid Metaheuristics 2005, Universitat Politècnica de Catalunya, Barcelona, Spain, August 29-30, 2005.

[81] H. Ishibuchi, T. Yoshida, T. Murata, (2003) "Balance between Genetic Search and Local Search in Memetic Algorithms for Multiobjective Permutation Flowshop Scheduling", IEEE Transactions on Evolutionary Computation, Vol. 7 , No. 2, pp. 204–233.

[82] V. Cherkassky, F. Mulier, (1998) Learning From Data: Concepts, Theory, and Methods, Wiley, New York, 1998.

[83] B. Everitt, S. Landau, M. Leese, (2001) Cluster Analysis, Arnold, London, 2001.

[84] A. Abraham, L. C. Jain, R. Goldberg, (2005) Evolutionary Multiobjective Optimization: Theoretical Advances and Applications, Springer Verlag, London.

[85] E. R. Hruschka, J. G. B. Campello, A. A. Freitas, "A Survey of Evolutionary Algorithms for Clustering", IEEE Transactions on Systems, Man, and Cybernetics - Part C: Applications and Reviews.

[86] K. Ramachandra Rao et.al, (2012) "Unsupervised Classification of Uncertain Data Objects in Spatial Databases Using Computational Geometry and Indexing Techniques", International Journal of Engineering Research and Applications, Vol. 2, No. 2, pp. 806-814. ISSN No. 2248-9622, http://www.ijera.com/papers/Vol2_issue2/ EG22806814.pdf.

[87] M. Sarkar, B. Yegnanarayana, D. Khemani, (1997) "A clustering algorithm using an evolutionary programming-based approach", Pattern Recognition Letters, Vol. 18, pp. 975-986.

[88] K. Karteeka Pavan, A.V. Dattatreya Rao, A. Appa Rao, (2011) "Robust seed selection algorithm for k-means type algorithms" , International Journal of Computer Science & Information Technology (IJCSIT), Vol. 3, No. 5, DOI: 10.5121/ijcsit.2011.3513.

[89] H. Dong, Y. Dong, C. Zhou, et.al, (2009) "Fuzzy clustering algorithm based on evolutionary programming", Expert Systems with Applications, doi:10.1016/j.eswa.2009.04.031.

[90] Arnold. Everitt, B. S., Landau, S., Leese, M., & Stahl, D. (2001) Hierarchical clustering. Cluster Analysis, 5th Edition, 71-110.

[91] L. Kaufman, P. Rousseeuw, (1990) Finding Groups in Data: An Introduction to Cluster Analysis, Wiley.

[92] D. Sankoff, J. Kruskal, (1999) "Time Warps, String Edits, and Macromolecules: The Theory and Practice of Sequence Comparison", CA, CSLI Publications.

[93] V. Guralnik, G. Karypis, (2001) "A scalable algorithm for clustering sequential data" in Proceedings of 1st IEEE Int. Conf. Data Mining (ICDM'01), pp. 179–186.

[94] D. Gusfield, (1997) Algorithms on Strings, Trees, and Sequences: Computer Science and Computational Biology, Cambridge, U.K, Cambridge Univ. Press.

[95] F. Höppner, F. Klawonn, R. Kruse, (1999) Fuzzy Cluster Analysis: Methods for Classification, Data Analysis, and Image Recognition, New York: Wiley.

[96] T. Kohonen, (1990) "The self-organizing map" Proc. IEEE, Vol. 78, No. 9, September, pp. 1464–1480.

[97]  A. Baraldi, E. Alpaydin, (2002) "Constructive feed forward ART clustering networks—Part I and II", IEEE Trans. Neural Network, Vol. 13, No. 3, May, pp. 645–677.

[98]  K. Müller, S. Mika, et.al. (2001) "An introduction to kernel-based learning algorithms" IEEE Trans. Neural Network, Vol. 12, No. 2, pp. 181–201.

[99]  V. Vapnik, (1998) Statistical Learning Theory, Wiley, New York.

[100] A. Jain, R. Dubes, (1988) Algorithms for Clustering Data. Englewood Cliffs, NJ: Prentice-Hall, 1988.

[101] R. Sun, C. Giles, (2000) "Sequence learning: Paradigms, algorithms, and applications" in Proceeding of LNAI 1828, Berlin, Germany.

[102] A. Garg, A. Mangla, N. Gupta, V. Bhatnagar, (2006) "PBIRCH: A Scalable Parallel Clustering algorithm for Incremental Data", Database Engineering and Applications Symposium, International, 2006.

[103] C.A. Coello, http://delta.cs.cinvestav.mx/~ccoello/EMOO, Cinvesav-ipn, Mexico.

## Authors

**Ramachandra Rao Kurada** is currently working as Asst. Prof. in Department of Computer Applications at Shri Vishnu Engineering College for Women, Bhimavaram. He has 12 years of teaching experience and is a part-time Research Scholar in Dept. of CSE, ANU, Guntur under the guidance of Dr. K Karteeka Pavan. He is a life member of ISTE and CSI. His research interests are Computational Intelligence, Data warehouse & Mining, Networking and Securities.

**Dr. K. Karteeka Pavan** has received her PhD in Computer Science & Engg. from Acharya Nagarjuna University in 2011. Earlier she had received her postgraduate degree in Computer Applications from Andhra University in 1996. She is having 16 years of teaching experience and currently she is working as Professor in Department of Information Technology of RVR &JC College of Engineering, Guntur. She has published more than 20 research publications in various International Journals and Conferences. Her research interest includes Soft Computing, Bioinformatics, Data mining, and Pattern Recognition.

**Dr. A.V.Dattatreya Rao** has received PhD in Statistics from Acharya Nagarjuna University in 1987. Currently he is Principal, Acharya Nagarjuna University, Guntur. Earlier he was Professor in Department of Statistics, Acharya Nagarjuna University. He has published more than 45 research publications in various National, Inter National Journals. He has successfully guided 3 Ph.Ds and 5 M.Phils. His research interest includes Estimation Theory, Statistical Pattern Recognition, Directional Data Analysis and Image Analysis. He is a life member of professional societies like Indian Statistical Association, Society for Development of Statistics, Andhra Pradesh Mathematical society and a life member in ISPS. He acted as a chairperson for various conferences.